\documentclass[sigconf,nonacm]{acmart}
\AtBeginDocument{%
  }

\usepackage{subfig}
\usepackage{multirow}
\usepackage{tikz}
\usepackage{pgfplots}
\pgfplotsset{compat=1.18}

\usepackage{balance}

\copyrightyear{2025}
\acmYear{2025}
\setcopyright{cc}
\setcctype{by}
\acmConference[SUMAC '25]{Proceedings of the 7th International Workshop on analySis, Understanding and proMotion of heritAge Contents}{October 27--28, 2025}{Dublin, Ireland}
\acmBooktitle{Proceedings of the 7th International Workshop on analySis, Understanding and proMotion of heritAge Contents (SUMAC '25), October 27--28, 2025, Dublin, Ireland}\acmDOI{10.1145/3746273.3760209}
\acmISBN{979-8-4007-2055-0/2025/10}

\begin{document}

\title[DGME-T]{DGME-T: Directional Grid Motion Encoding for Transformer-Based Historical Camera Movement Classification}


\author{Tingyu Lin}
\orcid{0009-0008-9825-686X}
\affiliation{%
  \institution{Computer Vision Lab, TU Wien}
  \city{Vienna}
  \country{Austria}}
\email{tylin@cvl.tuwien.ac.at}

\author{Armin Dadras}
\orcid{0000-0001-6474-7208}
\affiliation{%
  \institution{Media Computing Group, UAS St. Pölten}
  \city{St. Pölten}
  \country{Austria}
}
\affiliation{%
  \institution{Computer Vision Lab, TU Wien}
  \city{Vienna}
  \country{Austria}
}

\author{Florian Kleber}
\orcid{0000-0001-8351-5066}
\affiliation{%
  \institution{Computer Vision Lab, TU Wien}
  \city{Vienna}
  \country{Austria}}

\author{Robert Sablatnig}
\orcid{0000-0003-4195-1593}
\affiliation{%
  \institution{Computer Vision Lab, TU Wien}
  \city{Vienna}
  \country{Austria}}


\renewcommand{\shortauthors}{Tingyu Lin, Armin Dadras, Florian Kleber, \& Robert Sablatnig}


\begin{abstract}
Camera movement classification (CMC) models trained on contemporary, high-quality footage often degrade when applied to archival film, where noise, missing frames, and low contrast obscure motion cues. We bridge this gap by assembling a unified benchmark that consolidates two modern corpora into four canonical classes and restructures the HISTORIAN collection into five balanced categories. Building on this benchmark, we introduce \textbf{DGME-T}, a lightweight extension to the Video Swin Transformer that injects directional grid motion encoding, derived from optical flow, via a learnable and normalised late-fusion layer. DGME-T raises the backbone’s top-1 accuracy from \textbf{81.78\,\% to 86.14\,\%} and its macro~$F_{1}$ from \textbf{82.08\,\% to 87.81\,\%} on modern clips, while still improving the demanding World-War-II footage from \textbf{83.43\,\% to 84.62\,\%} accuracy and from \textbf{81.72\,\% to 82.63\,\%} macro~$F_{1}$. A cross-domain study further shows that an intermediate fine-tuning stage on modern data increases historical performance by more than five percentage points. These results demonstrate that structured motion priors and transformer representations are complementary and that even a small, carefully calibrated motion head can substantially enhance robustness in degraded film analysis. Related resources are available at \url{https://github.com/linty5/DGME-T}.
\end{abstract}

\ccsdesc[500]{Computing methodologies~Computer vision tasks}
\ccsdesc[300]{Information systems~Multimedia information systems}
\ccsdesc[300]{Applied computing~Media arts}

\keywords{Camera Movement Classification, Historical Video, Optical Flow, Motion Encoding, Video Transformer, Domain Adaptation}




\begin{teaserfigure}
  \centering
  \includegraphics[width=0.9\linewidth]{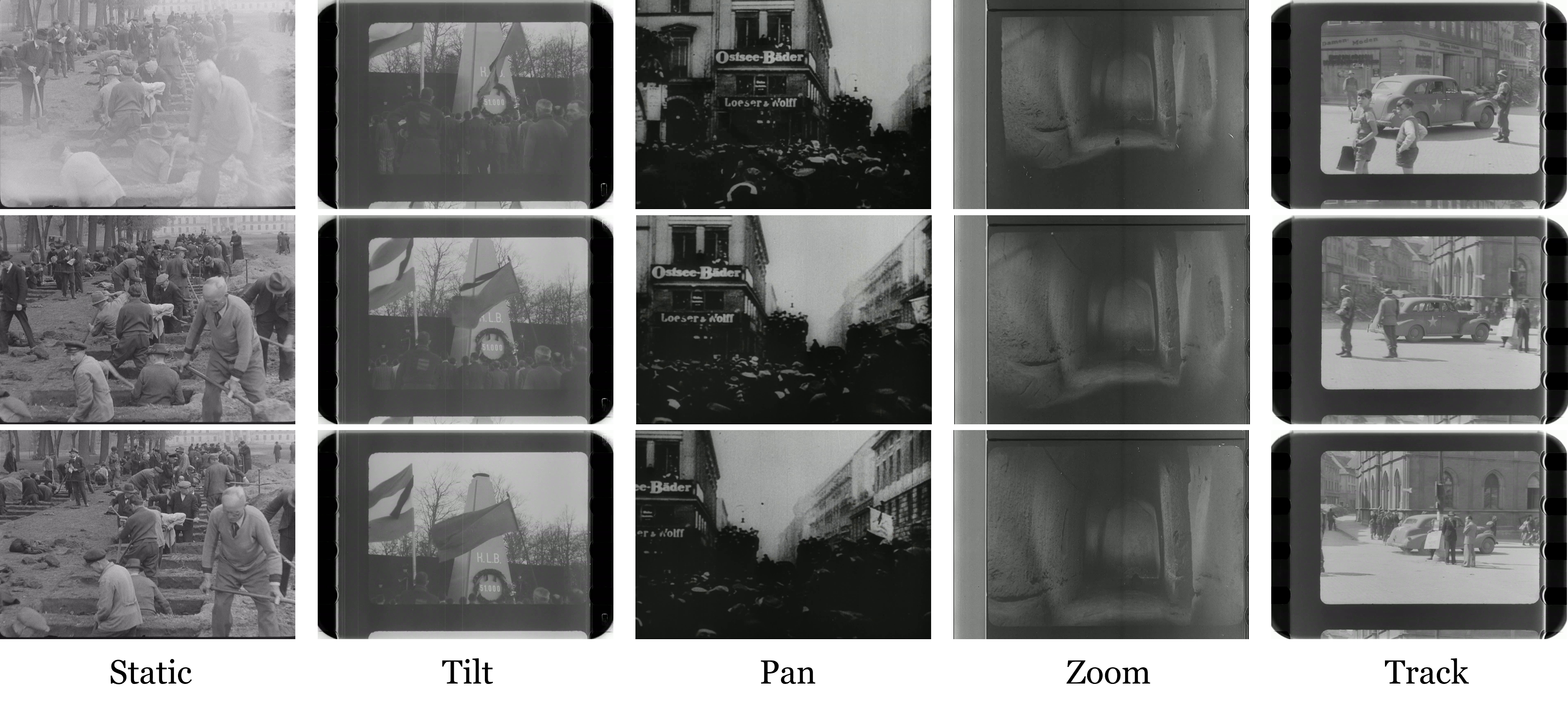}
  \caption{Example frames from the HISTORIAN dataset illustrating typical visual degradation, blur, and low contrast encountered in archival footage.}
  \label{fig:historian_teaser}
\end{teaserfigure}


\maketitle

\section{Introduction}

Camera movement plays a fundamental role in cinematic expression, shaping narrative comprehension, visual storytelling, and audience engagement \cite{bordwell1997history, bordwell2010film}. Recognizing and classifying such movements, known as Camera Movement Classification (CMC), involves assigning semantic labels such as \textit{pan}, \textit{tilt}, \textit{track}, \textit{dolly}, \textit{truck}, and \textit{zoom} to short video segments. Accurate CMC supports various applications in video analysis and film studies. Its importance becomes even more pronounced in the historical domain: systematic motion analysis provides film scholars with quantitative tools to study stylistic conventions \cite{bordwell1997history}. At the same time, cultural heritage institutions can leverage automated annotations to enrich metadata during digitization and cataloguing, thereby improving retrieval and curation of archival collections \cite{helm2022historian, lin2024enhancing}. Reliable motion labels also benefit restoration workflows and downstream tasks such as shot detection, summarization, and stylistic analysis \cite{rao2020unified}. These applications highlight that historical CMC is a technical challenge and a key enabler for scalable access to and preservation of visual heritage.

Traditionally, research on CMC has followed two primary trajectories. Initial approaches relied on handcrafted motion descriptors derived from macroblock motion vectors or optical flow fields \cite{hasan2014camhid, prasertsakul2017video}. While such approaches effectively capture coarse-grained motion patterns, they often struggle under unconstrained conditions or in complex camera movements. With the rapid advancement of deep learning, recent efforts have adopted convolutional neural networks (CNNs), recurrent networks (RNNs), and, more recently, Transformer-based architectures, demonstrating considerable success on modern video datasets \cite{chen2021ro, li2023lightweight, rao2020unified}. These data-driven methods learn discriminative features directly from visual input and generally outperform traditional descriptors due to their robust feature extraction capabilities.

However, despite impressive progress in modern datasets, applying existing CMC techniques to archival material remains an underexplored and significantly challenging problem. Historical imagery is often subject to severe degradations such as noise, blur, and contrast loss \cite{Zhao_2021_WACV}. When moving from still images to video, these degradations are further compounded by temporal inconsistencies: historical films, particularly wartime documentaries, exhibit unstable frame rates, exposure variations, and artifacts introduced during digitization (see Fig.~\ref{fig:historian_teaser}). Such characteristics substantially violate the assumptions inherent in modern video processing, namely the availability of clean, high-resolution imagery and smooth, predictable camera trajectories. Consequently, models trained on contemporary video datasets typically exhibit poor generalization when directly applied to historical footage. Furthermore, limited annotated historical datasets and the inherent difficulty of manually labeling degraded archival material exacerbate this challenge.

\begin{figure*}[htbp]
  \centering
  \includegraphics[width=0.9\linewidth ]{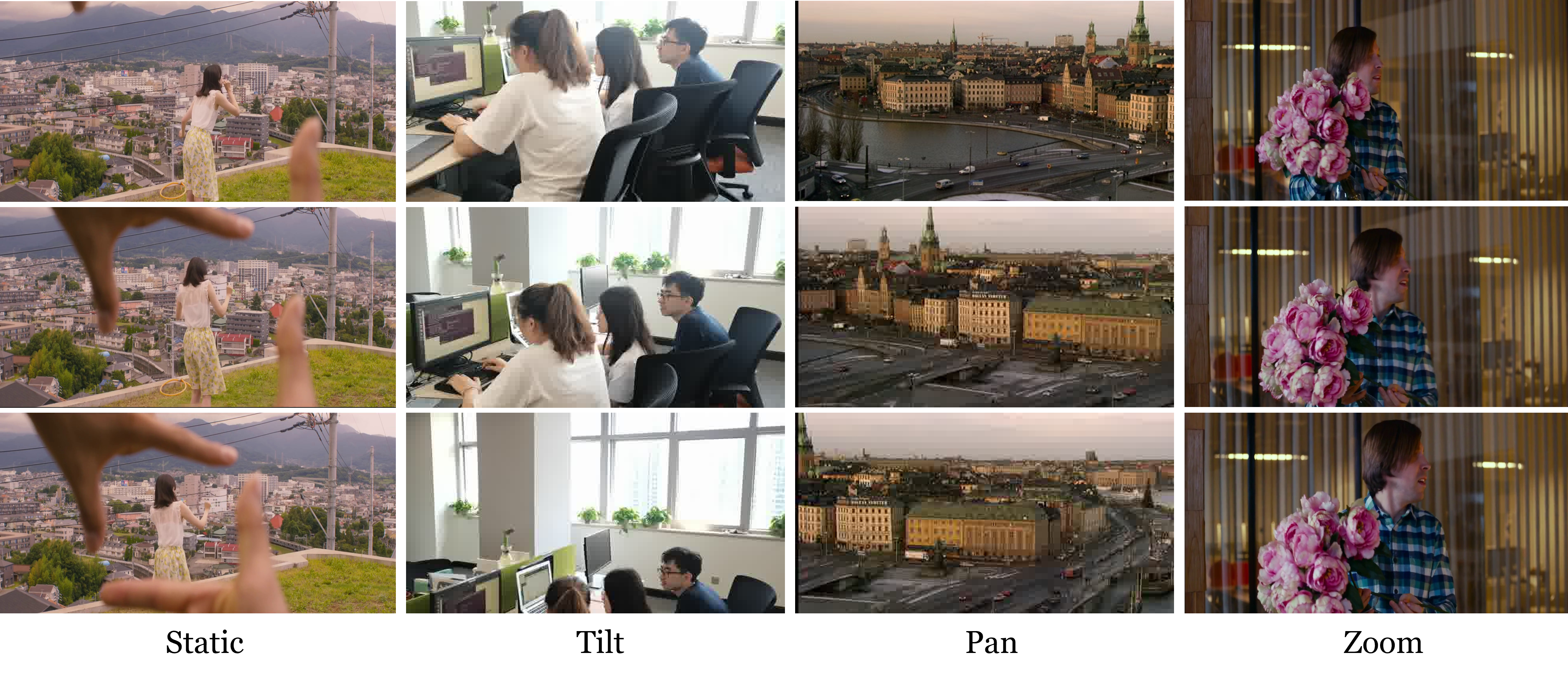}
  \caption{Example frames from the modern training dataset showing clean, high–resolution video content.}
  \label{fig:modern_sample}
\end{figure*}

Motivated by these challenges, this work systematically explores CMC specifically tailored to historical footage. We start by revisiting and unifying existing modern datasets, including MovieNet and MOVE-SET, to construct a balanced and comprehensive pre-training corpus comprising four categories: \textit{static}, \textit{tilt}, \textit{pan}, and \textit{zoom}. Representative samples of these four categories from the modern corpus are illustrated in Fig.~\ref{fig:modern_sample}. Additionally, we carefully adapt the HISTORIAN dataset, a dedicated collection of expertly annotated World War II archival films, by redefining ambiguous or underrepresented labels into a coherent five-category schema: \textit{static}, \textit{tilt}, \textit{pan}, \textit{zoom}, and \textit{track}. Example frames of these five categories from HISTORIAN are shown in Fig.~\ref{fig:historian_teaser}. This structured alignment of historical and modern datasets allows us to perform rigorous cross-domain evaluation and fine-tuning.

Transformer-based architectures have recently demonstrated strong capabilities in modeling long-range dependencies and subtle visual cues, making them effective for fine-grained camera movement classification \cite{arnab2021vivit,bertasius2021space,fan2021multiscale,liu2022video}. Nevertheless, prior studies have shown that Transformers without explicit temporal modeling or motion-sensitive mechanisms struggle on benchmarks that require capturing fine-grained movement cues \cite{arnab2021vivit,bertasius2021space}. This limitation becomes particularly critical in historical footage, where degradations and temporal inconsistencies demand robustness to low-level directional motion patterns. 

To address this, we propose Directional Grid Motion Encoding for Transformers (DGME-T), which augments a Transformer backbone with handcrafted directional motion cues integrated through learnable parameters and feature normalization, enhancing robustness to domain shifts and visual degradation. Extensive experiments confirm that DGME-T consistently outperforms baseline Transformers on modern datasets and achieves competitive or superior performance on the challenging HISTORIAN benchmark. In particular, it excels in recognizing static conditions, a class previously difficult due to subtle motion cues. Confusion matrix analyses further illustrate the approach's strengths and remaining limitations.

In summary, this work makes three main contributions. First, we present a unified framework for training and evaluating camera movement classifiers across modern and historical video datasets, facilitating robust cross-domain model transfer. Second, we introduce DGME-T, a lightweight integration of directional motion encoding with Transformer-based architectures that substantially improves CMC accuracy on modern datasets while effectively mitigating domain shifts in historical footage. Third, we conduct comprehensive comparative evaluations across modern and historical datasets, demonstrating the proposed approach's effectiveness and adaptability. 

The remainder of this paper is structured as follows. Section~\ref{sec:related} reviews related work, covering handcrafted descriptors, deep-learning-based approaches, and available datasets. Section~\ref{sec:method} presents the proposed DGME-T methodology in detail. Section~\ref{sec:dataset} outlines the dataset construction and label redefinition processes. Section~\ref{sec:experiments} reports comprehensive experimental results and analyses, including error analysis. Finally, Section~\ref{sec:conclusion} concludes with discussions of key findings and future directions.

\section{Related Work}\label{sec:related}

Research on CMC spans three key aspects. First, handcrafted motion descriptors explicitly encode statistics from optical flow or macroblock vectors. Second, data–driven deep learning methods leverage CNNs, RNNs, or Transformers to learn discriminative features directly from video, and also include adaptations of generic video classification backbones originally designed for action recognition. Finally, several dedicated datasets provide annotated material across modern and historical domains, forming the basis for training and evaluation. We briefly review each of these aspects below.

\textbf{Handcrafted descriptors.}
Early work used explicit motion statistics computed from optical flow or macroblock vectors. Wang and Cheong \cite{wang2009taxonomy} introduced a semantically guided taxonomy based on motion entropy and attention maps. Hasan et al.\ \cite{hasan2014camhid} proposed CAMHID, which builds histograms of macroblock vectors and classifies four movement types with an SVM. Prasertsakul et al.\ \cite{prasertsakul2017video} extended this idea by matching two-dimensional flow magnitude and orientation histograms to distinguish ten movements. Although efficient and interpretable, these methods assume a relatively clean video with simple background dynamics. In practice, they are easily disrupted by noise, grain, and irregular object motion, especially prevalent in degraded historical footage.

\textbf{Deep learning approaches.}
Several works design architectures explicitly for camera movement analysis. SGNet \cite{rao2020unified} fuses RGB, saliency, and segmentation cues to classify four coarse movements. MUL-MOVE-Net \cite{chen2021ro} combines CNNs with BiLSTMs to recognise nine directional and rotational motions, while Petrogianni et al.\ \cite{petrogianni2022film} incorporate low-level motion statistics within hybrid CNN/LSTM backbones. Li et al.\ \cite{li2023lightweight} propose LWSRNet, a lightweight 3D CNN that integrates multiple modalities and achieves strong accuracy on contemporary video.

Beyond task-specific designs, generic video recognition architectures have been widely applied to camera-motion understanding. Representative convolutional backbones include C3D \cite{Tran_2015_ICCV} and I3D \cite{Carreira_2017_CVPR} for complete 3D spatiotemporal modeling, R(2+1)D \cite{Tran_2018_CVPR}, which factorises spatial and temporal kernels, and TSN  \cite{wang2016temporal}, which aggregates sparsely sampled 2D features over long clips. Transformer-based models extend attention to video, such as Video Swin \cite{liu2022video}, TimeSformer \cite{bertasius2021space}, ViViT \cite{arnab2021vivit}, and MViT \cite{fan2021multiscale}, while SlowFast \cite{feichtenhofer2019slowfast}, S3D-G \cite{xie2018rethinking}, and MoViNets \cite{kondratyuk2021movinets} refine convolutional designs with multi-rate or mobile-efficient variants. Pretraining on large-scale benchmarks (e.g., Kinetics-400 \cite{kay2017kinetics}) is standard practice for these models and typically yields strong results on generic datasets such as UCF101 \cite{soomro2012ucf101}. However, their sensitivity to low-level directional motion cues under the degradations common in archival footage remains less explored, motivating approaches that complement high-level representations with explicit motion priors.

\textbf{Datasets.}
MovieShots \cite{rao2020unified} provides 46,857 annotated trailer shots spanning four broad movements, whereas MOVE-SET \cite{chen2021ro} offers over 100,000 frame pairs covering nine detailed motions. The Petrogianni corpus \cite{petrogianni2022film} includes 1,803 shots from feature films with ten nuanced categories. HISTORIAN \cite{helm2022historian} focuses on archival World War II material, annotating 838 segments with eight movement labels that include subtle classes such as \emph{track} and \emph{pedestal}. Visual quality, frame rate, and label granularity differ markedly across these datasets, hampering cross-domain evaluation.

Despite these advances, existing approaches still face notable limitations. Handcrafted descriptors provide interpretable motion cues but degrade severely under noise and unconstrained conditions. At the same time, deep learning models and generic video backbones capture richer semantics yet often remain insensitive to subtle directional motion patterns. In addition, variations in visual quality and label definitions across datasets hinder systematic comparison. These observations motivate the need for approaches that jointly exploit robust motion cues and high-level representations, supported by unified benchmarks spanning modern and historical footage.

\section{Methodology}\label{sec:method}

\begin{figure}[htbp]
  \centering
  \includegraphics[width=\linewidth]{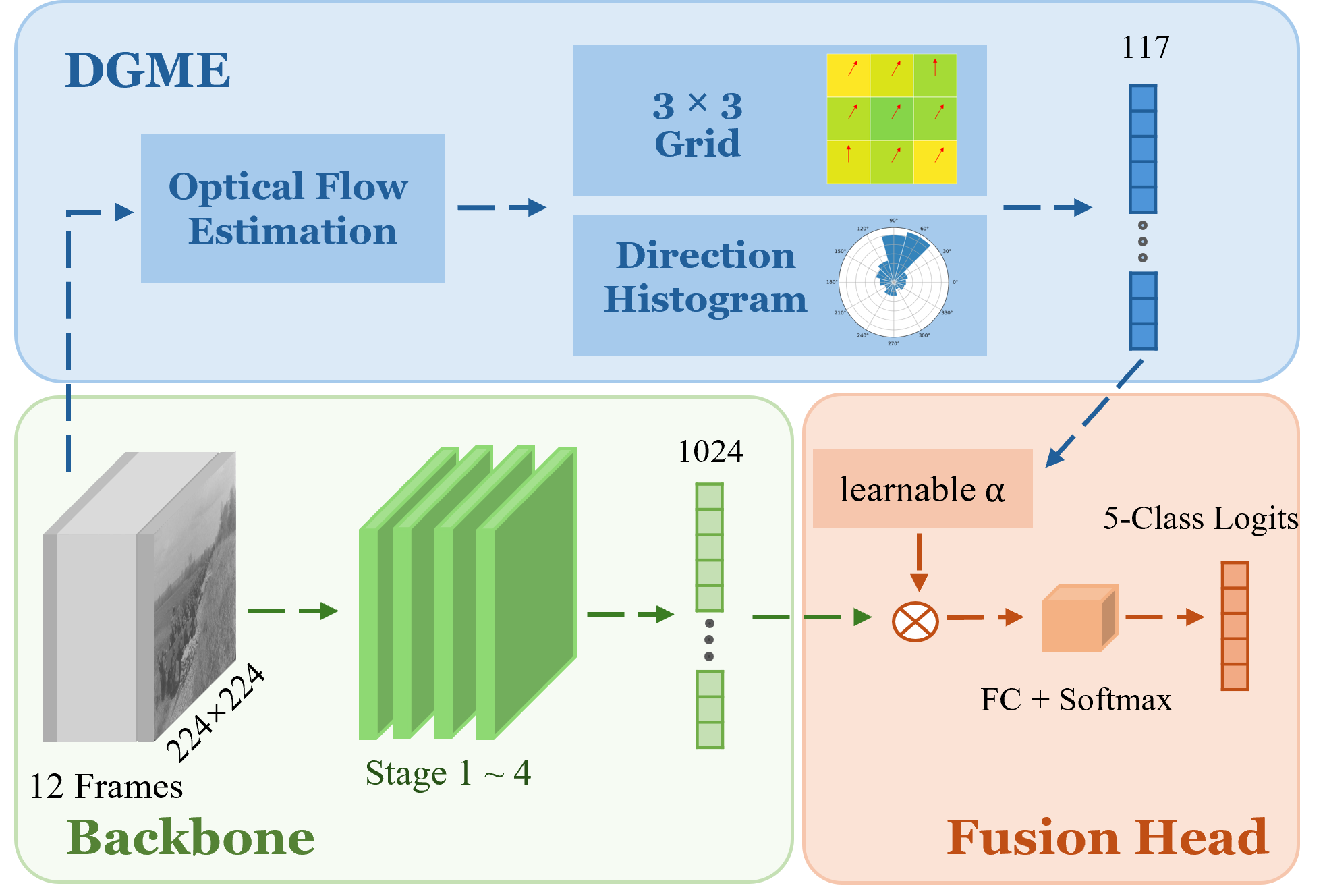}
  \caption{Overall architecture of DGME-T, combining directional motion encoding with a Video Swin Transformer backbone.}

  \label{fig:dgme_overview}
\end{figure}

In this section, we introduce our proposed method, DGME-T, designed to effectively classify camera movements, particularly addressing the unique challenges posed by historical video data. Our approach integrates directional motion information derived from optical flow with the powerful contextual modeling capabilities of Video Swin Transformer \cite{liu2022video}. To give an intuitive overview before delving into technical details, Fig.~\ref{fig:dgme_overview} illustrates how DGME and the Video Swin Transformer operate in parallel and are fused through a learnable head to produce five-class predictions.

\subsection{Directional Grid Motion Encoding}

Traditional handcrafted methods for CMC rely on extracting motion information explicitly from optical flow fields \cite{hasan2014camhid, prasertsakul2017video}. Inspired by these methods, DGME captures localized directional motion patterns using optical flow vectors computed by the Farneback algorithm \cite{farneback2003two}. Given consecutive frames from a video clip, we first compute the optical flow field, which yields horizontal and vertical motion components $(u, v)$ for each pixel. We convert these components into magnitude and angle representations as follows:
\begin{equation}
(m, \theta) = \text{cart2polar}(u, v),
\end{equation}
where $m$ denotes the magnitude and $\theta$ represents the angle in degrees. To reduce the effect of noise and minor irrelevant motions, we apply a threshold to the magnitude, retaining only motion vectors that exceed a specified threshold $m_{thr}$.

Next, the frame is spatially divided into a fixed $3\times3$ grid, and within each grid cell, we compute a weighted histogram of angles across predefined bins (e.g., 12 directional bins equally spaced from 0° to 360°). The weighting of each bin is proportional to the corresponding flow magnitudes, enabling emphasis on stronger, more relevant motion cues. Additionally, we include an extra "static" bin representing negligible movement. The histogram $h_{i,j}$ for grid cell $(i,j)$ is given by:

\begin{equation}
h_{i,j}(k) = \sum_{p \in \Omega_{i,j}} m(p) \cdot \mathbb{I}_{\theta(p) \in \text{bin}_k}, \quad k=1, \dots, K
\end{equation}

Where $\Omega_{i,j}$ denotes the set of pixels within grid cell $(i,j)$, $\mathbb{I}[\cdot]$ is the indicator function, and $K$ is the number of directional bins plus the static bin. All histograms are concatenated and L2-normalized to form a robust feature vector describing local directional motion patterns of the video segment.

\begin{figure*}[t]
\centering
\includegraphics[width=0.9\linewidth]{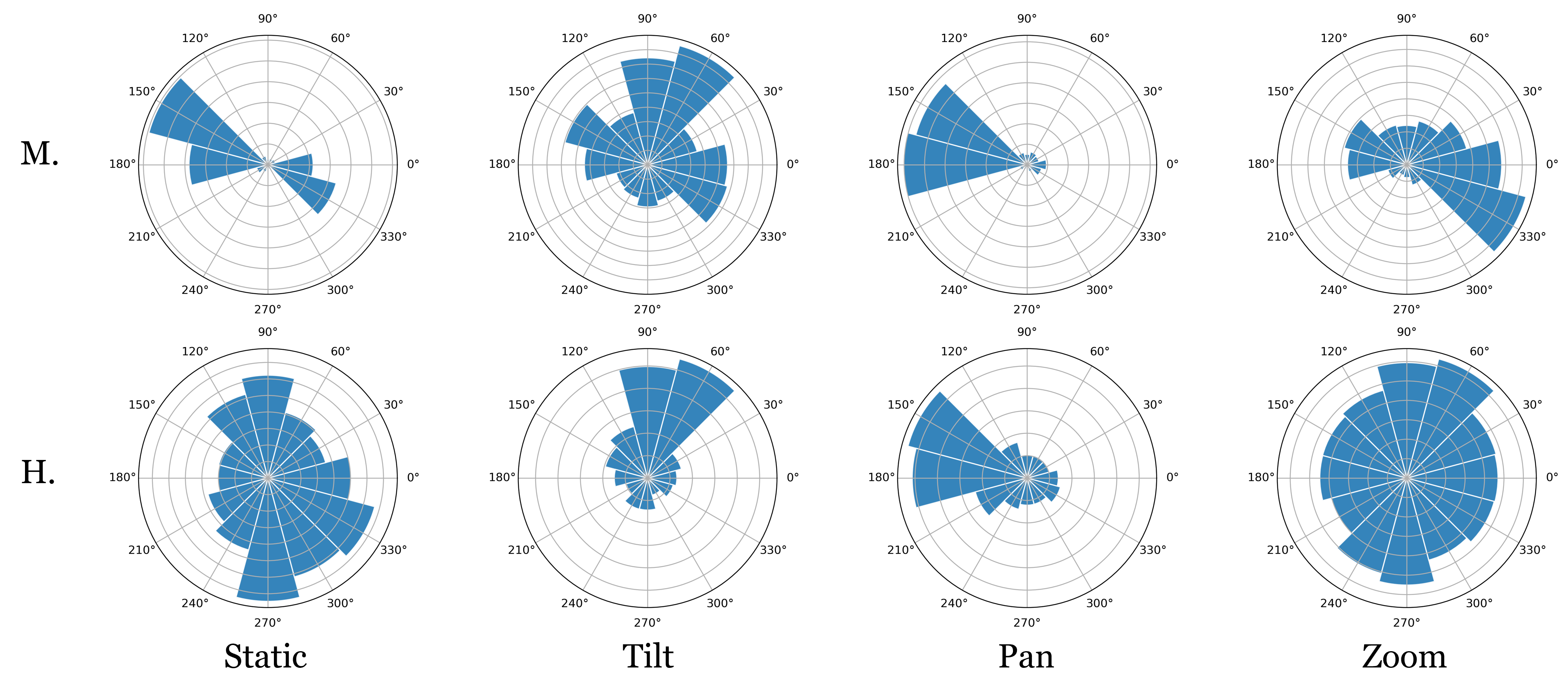}
\caption{Global $12$-direction rose diagrams for four movement classes, shown for the modern dataset (top row, see Fig.~\ref{fig:modern_sample}) and HISTORIAN (bottom row, see Fig.~\ref{fig:historian_teaser}).}
\label{fig:dgme_vis_polar}
\end{figure*}

To visualise what DGME captures, Fig.~\ref{fig:dgme_vis_polar} shows the $12$-bin directional histograms of four representative clips (\emph{static}, \emph{pan}, \emph{tilt}, \emph{zoom}) sampled from both the modern dataset and the HISTORIAN archive. A clean single peak characterises \emph{pan} and \emph{tilt}, whereas \emph{zoom} and cluttered \emph{static} sequences exhibit either a ring-like pattern or noisy, low-magnitude bars. 

\subsection{Integration with Video Swin Transformer}

Transformers have shown superior capability in modeling complex spatiotemporal relationships in video data \cite{liu2022video}, making them highly suitable for CMC tasks. Specifically, we utilize the Video Swin Transformer, which employs hierarchical self-attention blocks that effectively capture short- and long-range temporal dependencies.

However, Transformer models exhibit limited sensitivity to low-level directional motion cues, which are essential for reliable CMC, particularly in the presence of noise and degradations common in historical footage. We propose combining the DGME representation with the Transformer’s learned features at a late fusion stage to overcome this limitation.

Specifically, the Video Swin Transformer extracts a global spatiotemporal feature vector $F_{\text{swin}} \in \mathbb{R}^{C}$, where $C$ is the channel dimension after adaptive global pooling. We perform a late fusion by concatenating this global Transformer feature with the DGME representation $F_{\text{DGME}} \in \mathbb{R}^{D}$ as follows:
\begin{equation}
F_{\text{fusion}} = \left[ F_{\text{swin}},\ \alpha \cdot \text{LayerNorm}(F_{\text{DGME}}) \right]
\end{equation}

$\alpha$ is a learnable scalar parameter initialized at 1.0, enabling adaptive weighting of the DGME contribution, and $\text{LayerNorm}$ is applied solely to the DGME features to ensure scale consistency. This avoids DGME dominating the fused representation due to distributional differences across domains. The combined representation $F_{\text{fusion}}$ is then passed through a fully connected classification layer to produce the final class predictions:
\begin{equation}
\hat{y} = \text{softmax}(W_{f}F_{\text{fusion}} + b_{f})
\end{equation}
where $W_{f}$ and $b_{f}$ are trainable parameters of the fully connected layer.

\subsection{Feature Normalization and Domain Adaptation}

To address the domain gap between modern and historical datasets, we standardize the DGME features extracted from historical clips using statistics computed from the modern corpus. Specifically, given historical DGME features $F_{\text{hist}}$ and modern statistics (per-dimension mean $\mu_{\text{mod}}$ and standard deviation $\sigma_{\text{mod}}$), we apply z-score normalization:
\begin{equation}
F_{\text{hist}}^{\text{norm}} = \frac{F_{\text{hist}} - \mu_{\text{mod}}}{\sigma_{\text{mod}}}.
\end{equation}
Anchoring historical features to the modern scale ensures that motion cues degraded by noise, blur, or frame irregularities are interpreted on the same range as clean footage, rather than drifting toward a separate domain-specific representation. As later experiments confirm, this calibration is essential for stable transfer across domains.


\section{Dataset Construction and Label Redefinition}\label{sec:dataset}

We constructed and standardized datasets with coherent and balanced annotations to enable robust training and evaluation across modern and historical video domains. Directional Grid Motion Encoding (DGME) features were extracted using Farneback optical flow from uniformly sampled 12-frame video segments. To improve robustness, we applied standard preprocessing and augmentation procedures: frames were resized, cropped, and color-jittered during training, while evaluation used only resizing and center cropping for consistency. These preprocessing steps are kept consistent across modern and historical datasets, ensuring comparability of the extracted features. Next, we describe the construction of the modern and historical subsets and clarify the rationale behind our label definitions.

\subsection{Modern Dataset Construction}

The modern dataset was constructed by integrating relevant video segments from two publicly available datasets: MOVE-SET~\cite{chen2021ro} and MovieShots~\cite{rao2020unified}. These sources were selected due to their diverse yet complementary annotations. 

Originally, MOVE-SET contained various fine-grained camera movement labels, including descriptive terms like "stable," "up," "down," "left," "right," "in," and "out." To align with the target HISTORIAN categories, we redefined and aggregated these labels into four canonical classes: \textit{static}, \textit{pan}, \textit{tilt}, and \textit{zoom}. Specifically, segments labeled as "stable" were renamed as \textit{static}, "up" and "down" were grouped under \textit{tilt}, "left" and "right" were combined into \textit{pan}, and "in" and "out" were merged as \textit{zoom}. Similarly, MovieShots originally contained labels such as "static," "push," "pull," and "motion." To maintain consistency and clarity, we retained only the \textit{static} class and combined "push" and "pull" into the \textit{zoom} category, excluding the broadly defined "motion" class due to its ambiguity.

Due to significant imbalances in sample distribution across classes, we adopted a selective oversampling strategy. We increased the sample quantity for minority classes (tilt, pan, zoom) by repeating entries in the training annotation set. The validation set was not oversampled to ensure unbiased model evaluation. Figure~\ref{fig:modern_samples_bar} visualizes the final sample distribution of the modern dataset before and after oversampling. The video clips in the modern dataset were uniformly sampled at 12 frames per clip with a frame interval of 6. 

\definecolor{myorange}{RGB}{230,158,125}
\definecolor{myblue}{RGB}{20,112,176}
\definecolor{myyellow}{RGB}{241,210,120}

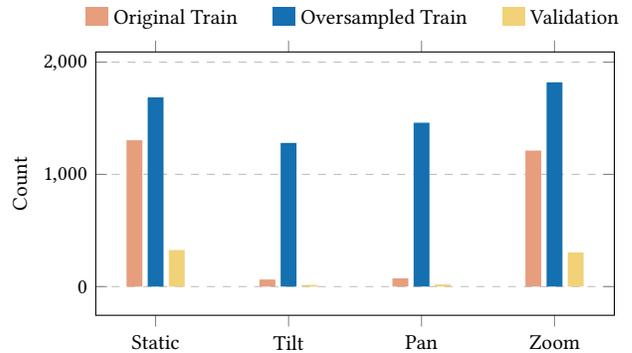
\begin{figure}[htbp]
\centering
\begin{tikzpicture}
\begin{axis}[
    ybar,
    bar width=6pt,
    width=\linewidth,
    height=0.6\linewidth,
    enlargelimits=0.15,
    ylabel={Count},
    symbolic x coords={Static,Tilt,Pan,Zoom},
    xtick=data,
    ymajorgrids=true,
    grid style=dashed,
    tick label style={/pgf/number format/fixed, font=\small},
    ylabel style={font=\small},
    xticklabel style={font=\small},
    legend style={
      at={(0.5,1.05)},
      anchor=south,
      legend columns=3,
      font=\small,
      /tikz/every even column/.append style={column sep=0.4cm},
      draw=none,
      fill=none
    },
    legend image code/.code={
      \draw[#1,draw=none] (0cm,-0.1cm) rectangle (0.3cm,0.15cm);
    }
]

\addplot[fill=myorange, draw=none] coordinates {(Static,1304) (Tilt,63) (Pan,73) (Zoom,1212)};
\addplot[fill=myblue, draw=none] coordinates {(Static,1686) (Tilt,1280) (Pan,1460) (Zoom,1820)};
\addplot[fill=myyellow, draw=none] coordinates {(Static,326) (Tilt,15) (Pan,19) (Zoom,304)};
\legend{Original Train, Oversampled Train, Validation}

\end{axis}
\end{tikzpicture}
\caption{Sample distribution of the modern dataset before and after oversampling.}
\label{fig:modern_samples_bar}
\end{figure}

\subsection{Historical Dataset Adaptation}

The HISTORIAN dataset~\cite{helm2022historian}, originally annotated with eight camera movement classes ("pan", "tilt", "zoom", "dolly", "truck", "track", "pedestal", "pan\_tilt"), consists of 838 segments from historical World War II archival footage. We redefined and merged ambiguous or underrepresented categories to ensure adequate training and comparison. Specifically, we combined visually similar movements—"truck" into \textit{pan}, "pedestal" into \textit{tilt}, "dolly" into \textit{zoom}—and excluded the "pan\_tilt" category due to insufficient sample size. Additionally, we introduced a clearly defined \textit{static} category. Moreover, we retained the \textit{track} category to test the model’s semantic understanding capability, despite it not existing in the modern dataset. Table~\ref{tab:historian_samples} presents the final category composition.

\begin{table}[htbp]
\centering
\small
\caption{Revised HISTORIAN dataset sample distribution.}
\begin{tabular}{lccccc}
\toprule
Class & Static & Tilt & Pan & Zoom & Track \\
\midrule
Source & new & tilt+pedestal & pan+truck & zoom+dolly & track \\
Count & 82 & 116 & 304 & 77 & 252 \\
\bottomrule
\end{tabular}
\label{tab:historian_samples}
\end{table}

We employed a class-balanced stratified split for training, validation, and testing, adopting a 6:2:2 ratio. Figure~\ref{fig:historian_splits_bar} visualizes the class-balanced data splits across train, validation, and test subsets.

\definecolor{myorange}{RGB}{230,158,125}
\definecolor{myblue}{RGB}{20,112,176}
\definecolor{myyellow}{RGB}{241,210,120}

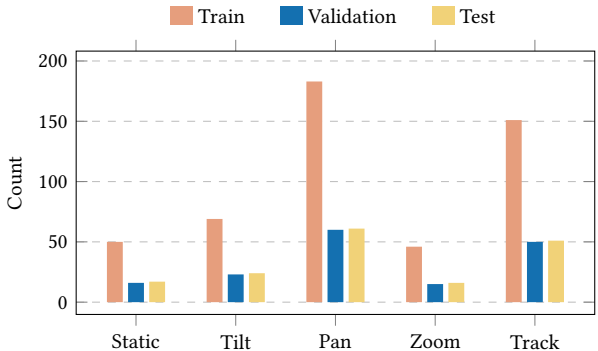
\begin{figure}[htbp]
\centering
\begin{tikzpicture}
\begin{axis}[
    ybar,
    bar width=6pt,
    width=\linewidth,
    height=0.6\linewidth,
    enlargelimits=0.15,
    ylabel={Count},
    symbolic x coords={Static,Tilt,Pan,Zoom,Track},
    xtick=data,
    ymajorgrids=true,
    grid style=dashed,
    tick label style={/pgf/number format/fixed, font=\small},
    ylabel style={font=\small},
    xticklabel style={font=\small},
    legend style={
      at={(0.5,1.05)},
      anchor=south,
      legend columns=3,
      font=\small,
      /tikz/every even column/.append style={column sep=0.4cm},
      draw=none,
      fill=none
    },
    legend image code/.code={
      \draw[#1,draw=none] (0cm,-0.1cm) rectangle (0.3cm,0.15cm);
    }
]

\addplot[fill=myorange, draw=none] coordinates {(Static,50) (Tilt,69) (Pan,183) (Zoom,46) (Track,151)};
\addplot[fill=myblue, draw=none] coordinates {(Static,16) (Tilt,23) (Pan,60) (Zoom,15) (Track,50)};
\addplot[fill=myyellow, draw=none] coordinates {(Static,17) (Tilt,24) (Pan,61) (Zoom,16) (Track,51)};
\legend{Train, Validation, Test}

\end{axis}
\end{tikzpicture}
\caption{Class-balanced splits for the HISTORIAN dataset across five categories.}
\label{fig:historian_splits_bar}
\end{figure}

To address domain discrepancies, we standardized HISTORIAN DGME features using the mean and standard deviation derived from the modern training dataset:
\begin{equation}
F_{\text{hist}}^{\text{norm}} = \frac{F_{\text{hist}} - \mu_{\text{mod}}}{\sigma_{\text{mod}}},
\end{equation}
where $\mu_{\text{mod}}$ and $\sigma_{\text{mod}}$ represent the mean and standard deviation computed across the modern dataset. The normalization was independently applied to train, validation, and test subsets.

\section{Experiments}\label{sec:experiments}

\begin{figure*}[htbp]
    \centering
    \subfloat[CAMHID -- Modern]{%
        \includegraphics[width=0.30\linewidth]{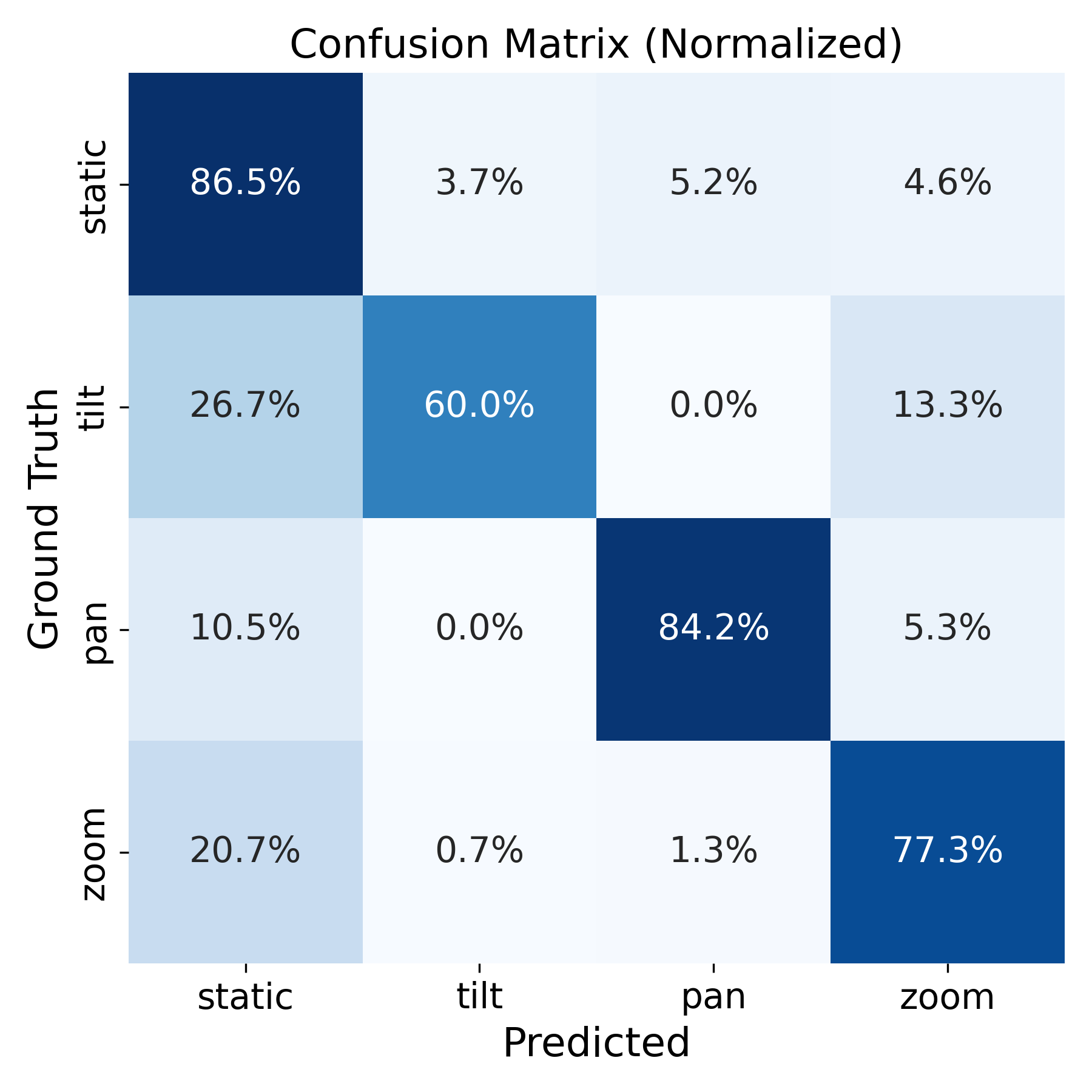}}
    \hfill
    \subfloat[Video Swin -- Modern]{%
        \includegraphics[width=0.30\linewidth]{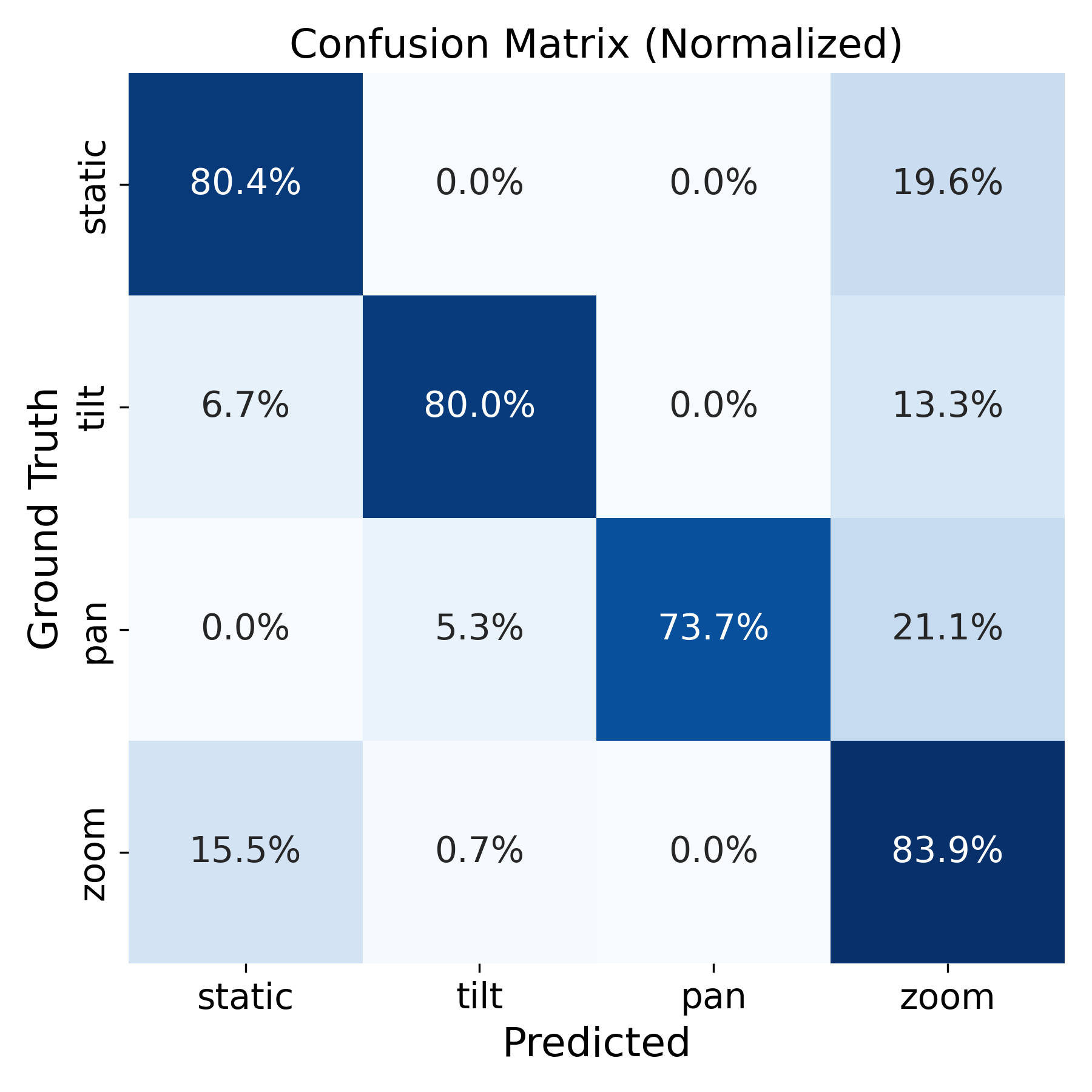}}
    \hfill
    \subfloat[DGME-T -- Modern]{%
        \includegraphics[width=0.37\linewidth]{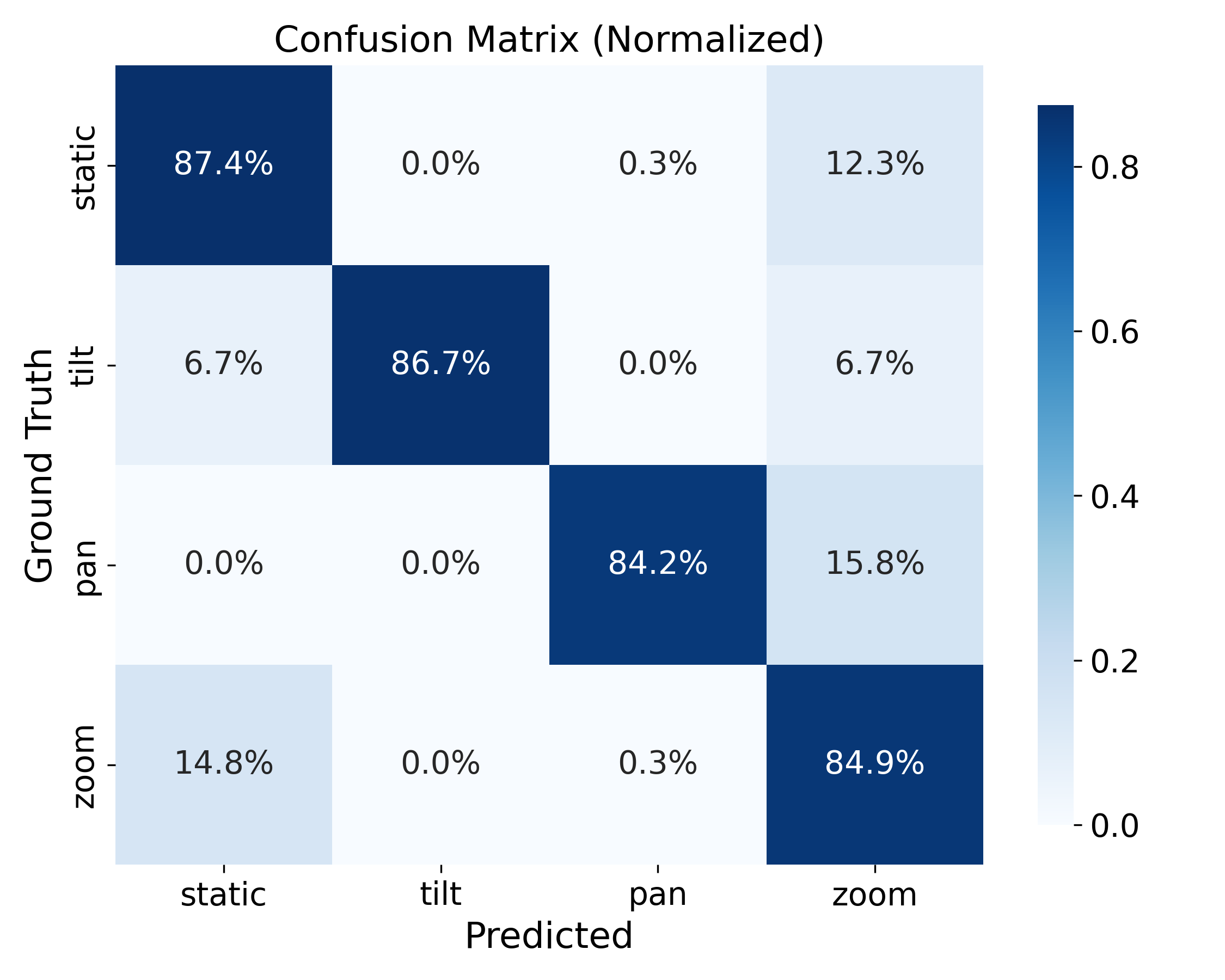}}\\[4pt]
    \subfloat[CAMHID -- HISTORIAN]{%
        \includegraphics[width=0.30\linewidth]{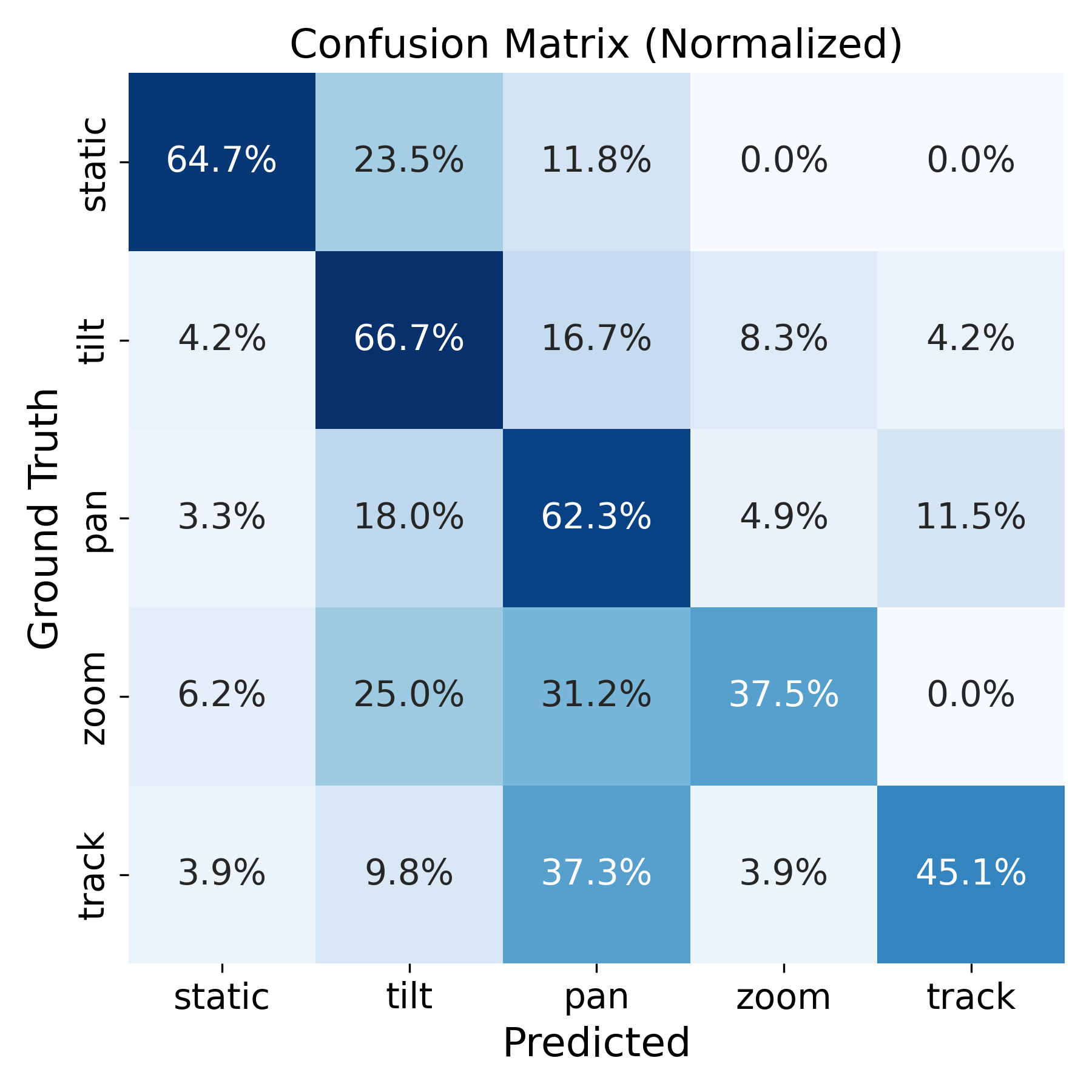}}
    \hfill
    \subfloat[Video Swin -- HISTORIAN]{%
        \includegraphics[width=0.30\linewidth]{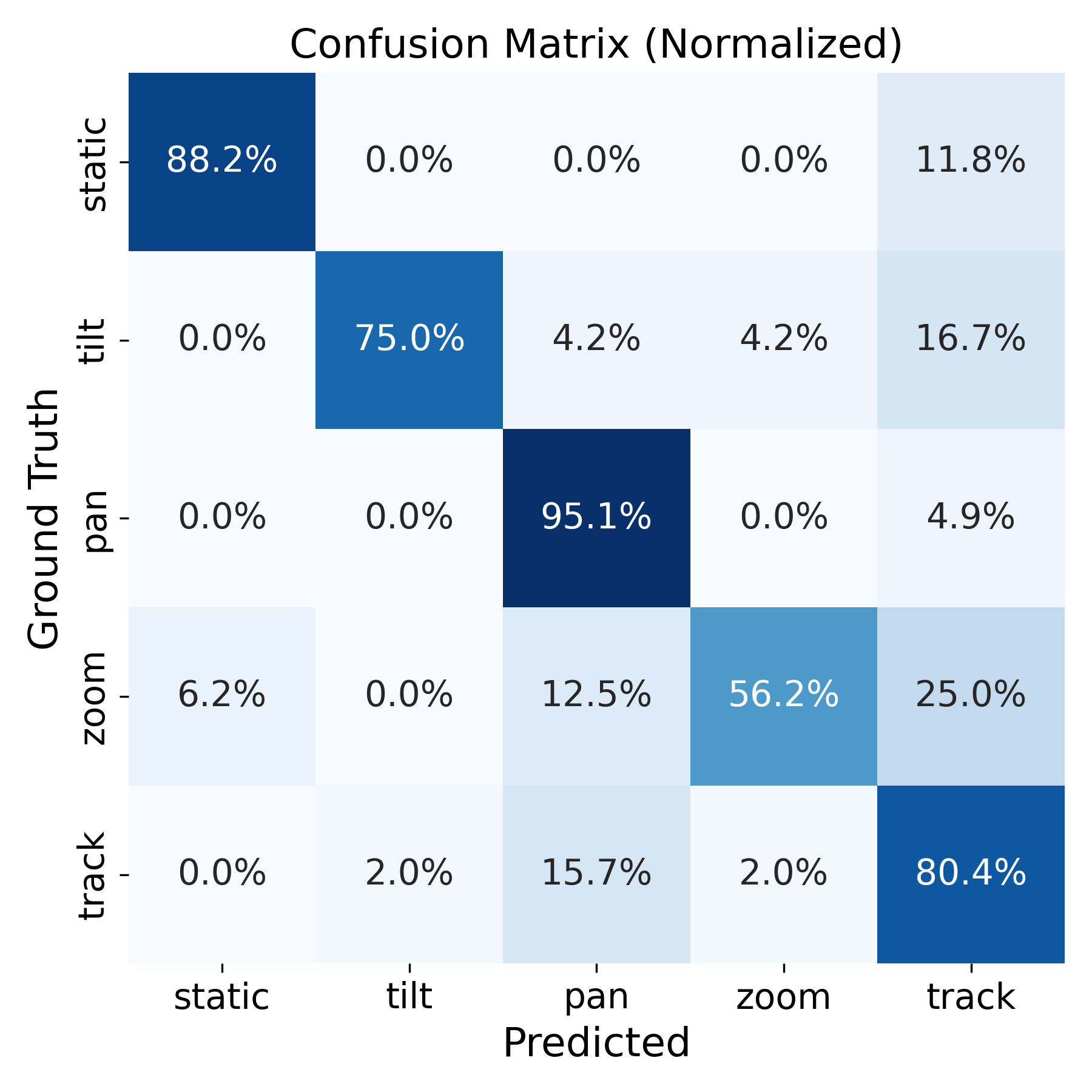}}
    \hfill
    \subfloat[DGME-T -- HISTORIAN]{%
        \includegraphics[width=0.37\linewidth]{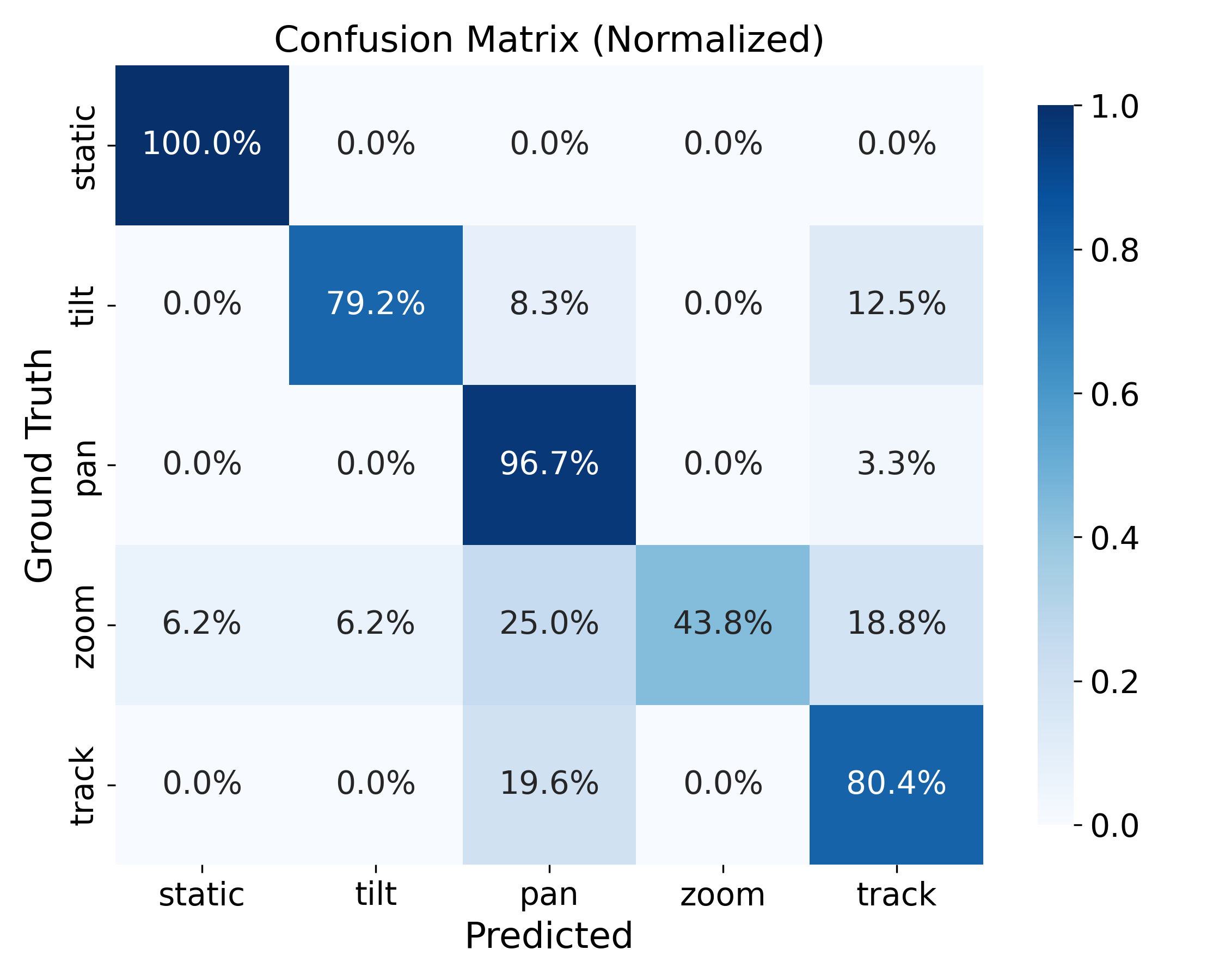}}
    \caption{Confusion matrices for three models on modern (top row) and HISTORIAN (bottom row) datasets.}
    \label{fig:cm_big}
\end{figure*}

We evaluate our proposed DGME-T method through comprehensive experiments on both modern and historical datasets. The backbone used for all deep learning models is Video Swin Transformer (Base variant), with an input clip length of 12 frames sampled every six frames. Frames are resized and center-cropped to $224\times224$ resolution. For training, we apply multi-scale cropping and color jittering to improve generalization. DGME features are fused with the final pooled token via late fusion, followed by a learnable scalar multiplier and LayerNorm. The models are trained using the AdamW optimizer with a cosine annealing scheduler over 12 epochs and early stopping. We report Top-1 Accuracy and Macro F1-score. All evaluations are performed on held-out validation or test sets described in Section~\ref{sec:dataset}, and representative confusion matrices are shown in Fig.~\ref{fig:cm_big}.

\subsection{Cross-Domain Transfer}

We investigate whether an intermediate pre-training stage on the modern corpus benefits final performance on HISTORIAN. We compare two variants of the \emph{Video Swin Transformer}. \textbf{Kinetics-only} is first pre-trained on Kinetics-400 for generic action recognition and then fine-tuned on HISTORIAN. \textbf{Modern-Historical} adds an extra fine-tuning step on the modern dataset before adapting to HISTORIAN. The detailed per-class results are given in Table~\ref{tab:cross_domain_class}, and the macro statistics are visualized in Figure~\ref{fig:cross_domain_macro_bar}.

\begin{table}[htbp]
\centering
\small
\caption{Per-class precision (P), recall (R) and $F_{1}$ on HISTORIAN.  All numbers are percentages.}
\begin{tabular}{l|ccc|ccc}
\toprule
\multirow{2}{*}{Class} & \multicolumn{3}{c|}{Kinetics-only} & \multicolumn{3}{c}{Modern-Historical} \\
 & P & R & $F_{1}$ & P & R & $F_{1}$ \\
\midrule
Static & 88.24 & 88.24 & 88.24 & 93.75 & 88.24 & 90.91 \\
Tilt   & 81.82 & 75.00 & 78.26 & 94.74 & 75.00 & 83.72 \\
Pan    & 75.68 & 91.80 & 82.96 & 84.06 & 95.08 & 89.23 \\
Zoom   & 85.71 & 37.50 & 52.17 & 81.82 & 56.25 & 66.67 \\
Track  & 75.51 & 72.55 & 74.00 & 75.93 & 80.39 & 78.10 \\
\midrule
Macro avg. & 81.39 & 73.02 & 75.13 & 86.06 & 78.99 & 81.72 \\
\bottomrule
\end{tabular}
\label{tab:cross_domain_class}
\end{table}


\definecolor{myorange}{RGB}{230,158,125}
\definecolor{myblue}{RGB}{20,112,176}

\begin{figure}[htbp]
\centering
\begin{tikzpicture}
\begin{axis}[
    ybar,
    bar width=10pt,
    width=\linewidth,
    height=0.55\linewidth,
    enlargelimits=0.15,
    ymin=0, ymax=100,
    ylabel={Score (\%)},
    symbolic x coords={Accuracy, Macro P, Macro R, Macro F1},
    xtick=data,
    ymajorgrids=true,
    grid style=dashed,
    tick label style={font=\small},
    ylabel style={font=\small},
    xticklabel style={font=\small},
    legend style={
        at={(0.5,1.05)},
        anchor=south,
        legend columns=2,
        font=\small,
        /tikz/every even column/.append style={column sep=0.4cm},
        draw=none,
        fill=none
    },
    legend image code/.code={
      \draw[#1,draw=none] (0cm,-0.1cm) rectangle (0.3cm,0.15cm);
    }
]

\addplot[
    fill=myorange,
    draw=none,
    bar shift=-6pt
] coordinates {
    (Accuracy,78.11) (Macro P,81.39) (Macro R,73.02) (Macro F1,75.13)
};

\addplot[
    fill=myblue,
    draw=none,
    bar shift=6pt
] coordinates {
    (Accuracy,83.43) (Macro P,86.06) (Macro R,78.99) (Macro F1,81.72)
};

\node[font=\footnotesize, anchor=south, xshift=-8pt] at (axis cs:Accuracy,78.11) {78.11};
\node[font=\footnotesize, anchor=south, xshift=-8pt] at (axis cs:Macro P,81.39) {81.39};
\node[font=\footnotesize, anchor=south, xshift=-8pt] at (axis cs:Macro R,73.02) {73.02};
\node[font=\footnotesize, anchor=south, xshift=-8pt] at (axis cs:Macro F1,75.13) {75.13};

\node[font=\footnotesize, anchor=south, xshift=8pt] at (axis cs:Accuracy,83.43) {83.43};
\node[font=\footnotesize, anchor=south, xshift=8pt] at (axis cs:Macro P,86.06) {86.06};
\node[font=\footnotesize, anchor=south, xshift=8pt] at (axis cs:Macro R,78.99) {78.99};
\node[font=\footnotesize, anchor=south, xshift=8pt] at (axis cs:Macro F1,81.72) {81.72};

\legend{Kinetics-only, Modern - Historical}
\end{axis}
\end{tikzpicture}
\caption{Macro-level performance comparison for cross-domain transfer.}
\label{fig:cross_domain_macro_bar}
\end{figure}
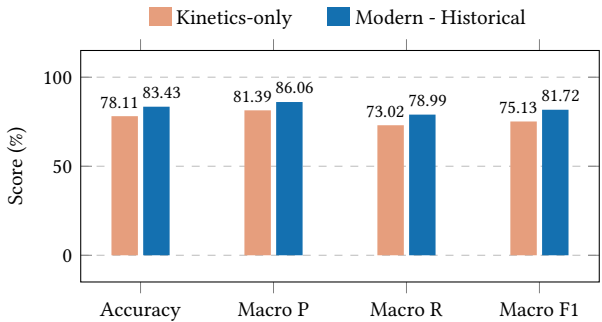

The additional modern pre-training stage consistently improves more than five percentage points in overall accuracy and over six points in macro $F_{1}$.  Gains are especially pronounced for \textit{tilt} and \textit{zoom}.  For \textit{tilt}, precision rises by thirteen points, while recall stays unchanged, indicating that appearance cues learned on modern footage help suppress false positives.  For the notoriously difficult \textit{zoom} class, $F_{1}$ increases from $52.2$ \% to $66.7$ \%, suggesting that the model better distinguishes subtle scale changes from camera translations once it has seen sufficient clean examples.  The improvement on \textit{pan} mainly manifests as higher recall, reflecting that temporally smooth lateral motion patterns in modern clips act as an effective prior for noisy archival sequences. 

In addition to staged pre-training, we also examine the role of feature calibration.  DGME statistics derived from optical flow exhibit different scales across modern and historical domains, owing to noise, contrast, and degradation variations.  Aligning these statistics through z-score normalization before fusion proves essential: on the HISTORIAN test set, DGME-T with normalization achieves 84.62\% accuracy and 82.63\% macro $F_{1}$, whereas removing this step reduces performance to 75.15\% accuracy and 72.63\% macro $F_{1}$.  A drop of ten percentage points in macro $F_{1}$ confirms that normalization is not a trivial preprocessing choice but a mechanism stabilizing the integration of handcrafted directional cues with Transformer features under domain shift. The study confirms that task-aligned source pre-training and careful domain calibration are crucial for transferring camera movement models to degraded archival footage.

\subsection{Model Comparison}

\begin{figure*}[htbp]
  \centering
  \includegraphics[width=0.9\linewidth]{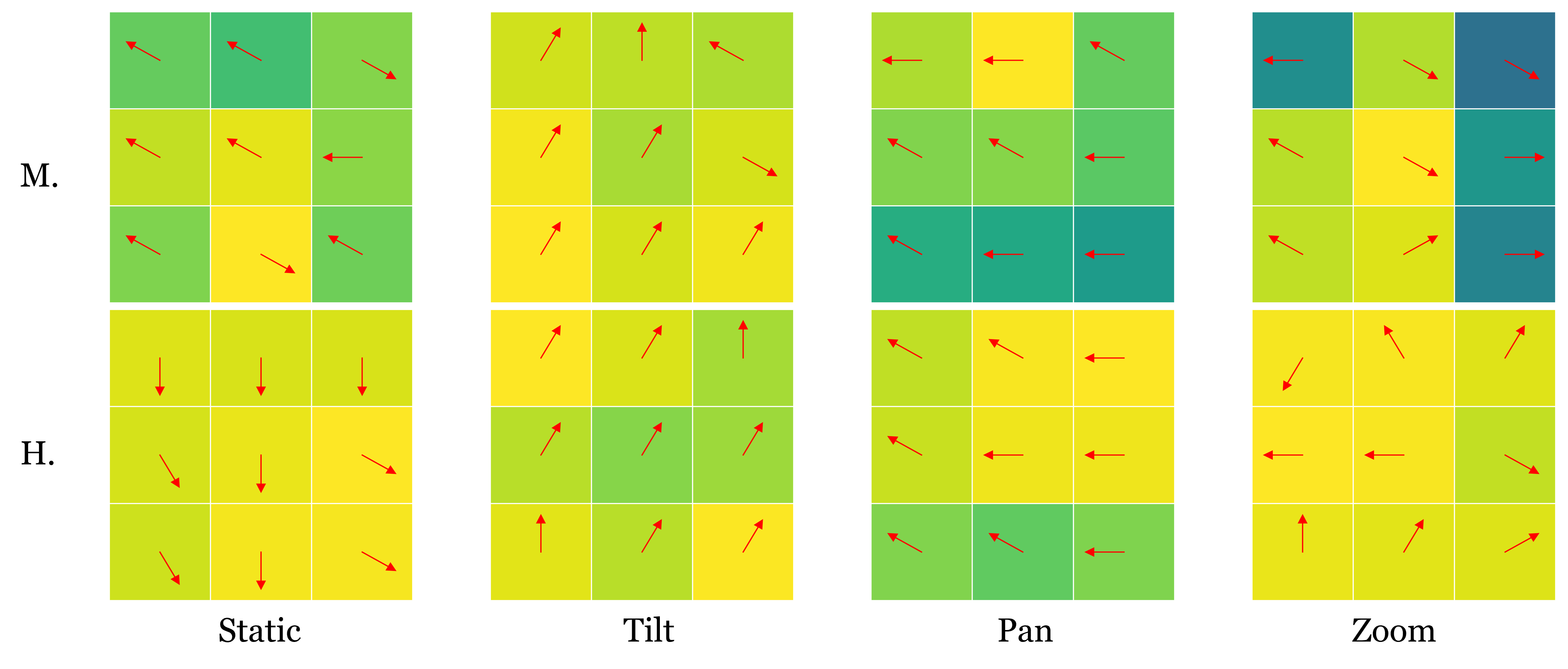}
  \caption{DGME 3\,$\times$\,3 grid visualisation for the same clips as Fig.~\ref{fig:dgme_vis_polar}. Cell colour encodes motion magnitude, arrows indicate the dominant direction in each cell.}
  \label{fig:dgme_vis_grid}
\end{figure*}

We compare three representative approaches on both domains: \emph{(i)~CAMHID}—a shallow classifier trained solely on DGME handcrafted features, \emph{(ii)~Video Swin Transformer} (deep baseline with no motion prior), and \emph{(iii)~DGME-T} (our hybrid late-fusion model). Figure~\ref{fig:cm_big} visualises class-wise confusion patterns, while Table~\ref{tab:model_comparison} reports overall accuracy and macro–$F_{1}$, with additional qualitative evidence provided in Fig.~\ref{fig:dgme_vis_grid}.

\begin{table}[htbp]
\centering
\small
\caption{Overall performance of three models on modern and historical datasets.}
\begin{tabular}{lcc|cc}
\toprule
 & \multicolumn{2}{c|}{Modern Dataset} & \multicolumn{2}{c}{HISTORIAN Dataset} \\
Model & Acc (\%) & $F_{1}$ (\%) & Acc (\%) & $F_{1}$ (\%) \\
\midrule
CAMHID (DGME-only) & 81.63 & 68.05 & 55.62 & 54.22 \\
Video Swin & 81.78 & 82.08 & 83.43 & 81.72 \\
DGME-T (Ours) & \textbf{86.14} & \textbf{87.81} & \textbf{84.62} & \textbf{82.63} \\
\bottomrule
\end{tabular}
\label{tab:model_comparison}
\end{table}

CAMHID shows an interesting contrast across domains. On the modern corpus, it attains respectable accuracy but a modest macro $F_{1}$: its histogram-based features cope well with the over-represented \textit{zoom} clips yet struggle with the long-tailed \textit{tilt} class, for which directional variance is subtle and sampling imbalance severe. When transferred to HISTORIAN, CAMHID’s advantage on \textit{zoom} disappears and its performance on \textit{track} collapses, underscoring that purely geometric flow cues lack the semantic sensitivity needed for object-following shots.

Video Swin Transformer is much more stable. Its appearance-driven representation secures strong results on both datasets and, in particular, yields high precision and recall on the semantically demanding \textit{track} class. Introducing DGME further lifts the model on the modern set, improving macro $F_{1}$ by 5.7 percentage points through reducing confusion between symmetric directions such as \textit{pan} and \textit{tilt}. In the historical domain, the hybrid gains are minor. DGME-T strengthens \textit{static}, \textit{pan}, and \textit{tilt}, with the \textit{static} category classified entirely correctly, demonstrating that low-magnitude directional priors are effective for deciding whether the camera is moving at all. Conversely, the extra descriptor offers little benefit for \textit{track} and slightly hurts \textit{zoom}, suggesting that flow noise and scale ambiguity outweigh the prior value in these cases. Even so, DGME-T still edges out the deep baseline by nearly one percentage point in accuracy and macro $F_{1}$ and remains ahead of CAMHID.

Qualitative inspection of Fig.~\ref{fig:dgme_vis_grid} supports the numerical trends in Table~\ref{tab:model_comparison}. For both datasets, DGME produces arrow fields that align with the expected motion patterns for \emph{pan} and \emph{tilt}, explaining the 5.7-point macro $F_{1}$ improvement obtained by DGME-T on the modern corpus. The descriptor also highlights the failure cases: (i)~\emph{static} frames contaminated by moving foreground (hands intruding from the border) mislead the purely motion-based CAMHID, and (ii)~for \emph{zoom}, when overall scaling of the frame coincides with substantial object or background changes, the resulting flow field lacks a clear dominant direction, limiting the benefit of DGME on HISTORIAN and explaining the smaller gain observed in Table~\ref{tab:model_comparison}. The visual evidence confirms that DGME supplies direction-sensitive priors complementary to the appearance-dominated Transformer backbone.

Overall, the study indicates that handcrafted motion encoding is a valuable complement rather than a standalone solution: it compensates for the Transformer’s weakness on movement direction, boosts performance in data-rich modern scenarios, and, despite mixed effects on individual classes, delivers a net gain in challenging archival footage. With better domain calibration or mid-level fusion strategies, we expect the motion prior to yield further improvements.

\section{Conclusion}\label{sec:conclusion}

We addressed camera movement classification (CMC) in historical footage, where visual degradation and noise limit the effectiveness of models trained on modern video. We established a unified benchmark to enable robust evaluation by consolidating two contemporary datasets into four movement classes and restructuring the eight HISTORIAN labels into five well-defined categories. On this foundation, we introduced DGME-T, a lightweight extension to the Video Swin Transformer that integrates directional grid motion features via late fusion with learnable scaling and feature normalisation. DGME-T improves accuracy from 81.78\% to 86.14\% and macro F\textsubscript{1} from 82.08\% to 87.81\% on modern data, while also lifting HISTORIAN accuracy from 83.43\% to 84.62\% and macro F\textsubscript{1} from 81.72\% to 82.63\%. Removing the z-score calibration reduces macro F\textsubscript{1} by ten points, underscoring the need for domain-specific normalisation. These results demonstrate that motion-sensitive priors remain valuable even with strong Transformer backbones, and the framework can be extended by exploring alternative flow estimators, fusion strategies, or integration points. While our historical evaluation centres on archival footage, future work could incorporate various sources across different cinematic periods.

\begin{acks}
This research was funded in whole or in part by the Austrian Science Fund (FWF) under project grant no. DFH 37-N: "Visual Heritage: Visual Analytics and Computer Vision Meet Cultural Heritage." For open access purposes, the author has applied a CC BY public copyright license to any author accepted manuscript version arising from this submission. The authors acknowledge TU Wien Bibliothek for financial support through its Open Access Funding Programme.
\end{acks}


\bibliographystyle{ACM-Reference-Format}
\balance
\bibliography{sample-base}










\end{document}